%% file: main.tex
\documentclass[letterpaper,10pt,conference]{ieeeconf}
\usepackage[OT1]{fontenc}

\usepackage{cite}
\usepackage{amsmath,amssymb,amsfonts}
\usepackage{graphicx}
\usepackage[percent]{overpic}
\usepackage{multirow}
\usepackage{booktabs}
\usepackage[ruled,vlined,linesnumbered]{algorithm2e}
\usepackage{textcomp}
\usepackage[table]{xcolor}
\usepackage[hidelinks]{hyperref}

\usepackage{xcolor}
\definecolor{coopgreen}{HTML}{367E75}
\definecolor{fixedred}{HTML}{C77676}
\definecolor{spacerose}{HTML}{C7ABAD}
\definecolor{feasiblegreen}{HTML}{AACFC0}
\definecolor{fiberpurple}{HTML}{9A87A8}

\usepackage{xcolor}
\newcommand{\pz}[1]{{#1}}

\def\BibTeX{{\rm B\kern-.05em{\sc i\kern-.025em b}\kern-.08em
    T\kern-.1667em\lower.7ex\hbox{E}\kern-.125emX}}

\begin{document}

\title{
\textbf{CAMP}: Cooperative Arm--Hand Motion Planning in Constrained Spaces
}

\author{
\authorblockN{
Ziyuan Wang$^{1,2}$,
Yunlong Shan$^{3}$,
Fei Mo$^{1,2}$,
Sichao Liu$^{4}$,
David Navarro-Alarcon$^{5}$,
Jia Pan$^{6}$, \\
Kosta Jovanovi\'c$^{7}$,
Xin Jiang$^{1,*}$,
Peng Zhou$^{2,*}$
}
\authorblockA{
$^{1}$Department of Mechanical Engineering and Automation, Harbin Institute of Technology, Shenzhen, China \\
$^{2}$School of Advanced Engineering, Great Bay University, Dongguan, Guangdong, China \\
$^{3}$State Key Laboratory for Multi-target Natural Medicine, China Pharmaceutical University, Nanjing, China \\
$^{4}$Department of Production Engineering, KTH Royal Institute of Technology, Stockholm, Sweden \\
$^{5}$Department of Mechanical Engineering, The Hong Kong Polytechnic University, Kowloon, Hong Kong SAR, China \\
$^{6}$School of Computing and Data Science, The University of Hong Kong, Hong Kong SAR, China \\
$^{7}$Department of Signals and Systems, School of Electrical Engineering, University of Belgrade, Belgrade, Serbia \\
$^{*}$Corresponding authors
}
}

\maketitle

\begin{abstract}
\pz{
Coordinated arm--hand motion planning is fundamental to dexterous robotic manipulation in complex and constrained environments. A straightforward solution is to decompose the problem into separate arm path planning and hand motion generation; however, this poses a dilemma: decomposition can miss feasible solutions that require coordinated arm--hand adaptation along the path. Alternatively, directly planning in the high-dimensional joint arm--hand configuration space captures such coupling but faces a substantially enlarged search space and nonconvex collision constraints. 
To characterize this coupling, we formulate feasible hand fibers that capture collision-free hand configurations for each arm configuration. Based on this formulation, we propose CAMP, a high-success and efficient cooperative arm--hand motion planner for constrained environments. CAMP constructs candidate trajectories through layered hand search with local arm relaxation, then compactly represents them using endpoint-preserving via-point movement primitives (VMPs) for coarse-to-fine joint optimization.
Across six constrained simulation tasks, CAMP achieves 84.2--98.5\% planning success, outperforming alternative planners with competitive efficiency. Ablation studies verify the contributions of arm relaxation, VMP representation, and coarse-to-fine optimization, while real-robot experiments demonstrate CAMP on constrained manipulation tasks.
The project website is available at \url{https://camp-armhand.github.io/}.
}

\end{abstract}

\input{Sections/1_introduction}

\input{Sections/3_formulation}
\input{Sections/4_methodology}
\input{Sections/5_experiment_setup}
\input{Sections/6_Experiments.tex}
\input{Sections/7_conclusion}

\bibliographystyle{ieeetr-etal}
\bibliography{references}

\end{document}

%% file: Sections/1_introduction.tex
\section{Introduction}
\pz{
Dexterous manipulation relies on the coordinated motion of the robot arm and hand~\cite{billard2019trends,wang2026world}. While the arm provides large-scale positioning and orientation, finger articulation reshapes the hand geometry to interact with objects and accommodate surrounding obstacles. This coordination becomes particularly important in confined environments, where the collision-free hand configurations available to the robot can change substantially as the arm moves(Fig.~\ref{fig:introduction_motivation}). Prior studies have also shown that exploiting arm--hand redundancy can improve motion quality and facilitate singularity, joint-limit, and collision avoidance~\cite{patel2023unified,bullock2013handcentric}. These observations motivate planning the arm and hand cooperatively throughout the motion.

Achieving such cooperation, however, presents what we call \textit{the Arm--Hand Planning Dilemma}: planning the arm and hand separately reduces the search dimension but can miss motions that require their configurations to change jointly. Direct search in the complete arm--hand configuration space preserves this coupling, but global exploration becomes difficult as the number of active joints increases. Moreover, a collision-free arm path does not necessarily admit a continuous collision-free hand motion between the prescribed endpoints. The arm route may therefore need to be adjusted while searching for feasible hand motion.

\begin{figure}[!t]
    \centering
    \setlength{\abovecaptionskip}{2pt}
    \includegraphics[width=\columnwidth]{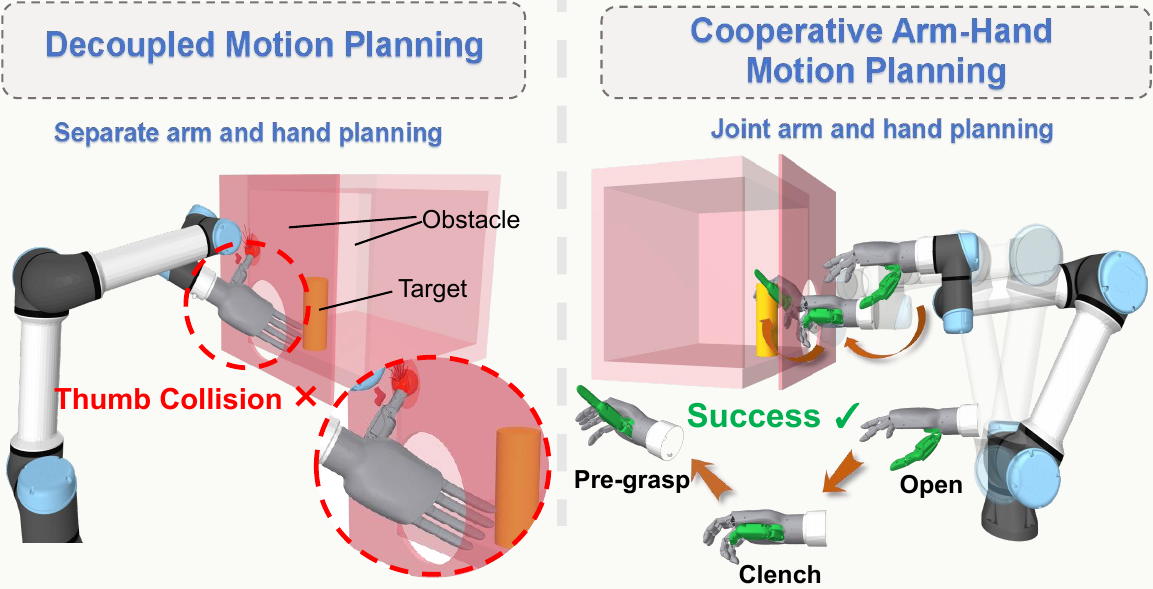}\par
    \nointerlineskip
    \caption{Motivation for cooperative arm--hand motion planning in a constrained opening. Decoupled planning leads to a thumb collision, whereas CAMP coordinates arm and hand motion to avoid it.}
    \label{fig:introduction_motivation}
    \vspace{-0.45\baselineskip}
\end{figure}

We interpret this arm--hand coupling through configuration-dependent feasible hand fibers~\cite{orthey2024multilevel,orthey2021sectionpatterns}. For each arm configuration, the corresponding fiber contains the hand configurations that keep the complete system free of environmental and self-collisions. As the arm moves, both the shape and connectivity of this feasible set may change. Consequently, fixing the hand configuration restricts the collision-free region available to the arm, while fixing an arm path restricts the feasible hand motions along that path. This perspective suggests that efficient cooperative planning should exploit the lower-dimensional arm space for global guidance without preventing local adjustment of either subsystem.

Existing methods address different aspects of arm--hand coordination. Dexterous grasp synthesis, such as BODex~\cite{chen2025bodex}, generates hand poses and finger configurations satisfying contact and grasp-quality objectives, while arm-aware grasp generation~\cite{jia2026armaware} additionally considers arm reachability and collision constraints. These methods primarily determine executable terminal configurations rather than planning collision-free trajectories. 
For motion planning, sampling-based planners~\cite{kuffner2000rrtconnect,gammell_bitstar_2020} provide global exploration but face increasing difficulty in high-dimensional spaces, whereas trajectory optimization methods~\cite{zucker2013chomp} efficiently refine continuous trajectories but depend strongly on initialization in nonconvex environments. 
QRRT and QRRT*~\cite{orthey2024multilevel} exploit simplified-space solutions to guide full-space search, but a path feasible for a simplified robot may not admit a collision-free lift for the complete arm--hand system. Section Patterns~\cite{orthey2021sectionpatterns} uses hand-designed local search patterns to lift a given base path, but lifting may fail when the guide leads toward a route that is infeasible for the complete robot or encounters difficulties beyond the recovery capabilities of these patterns.
Other arm--hand motion planning methods employ probabilistic roadmaps with hand-posture dimensionality reduction~\cite{rosell2011autonomous} or coarse-to-fine RRT search followed by path optimization~\cite{fan2025efficient}. However, the former requires captured human postures and hand-specific mapping, whereas the latter considers only arm motion in its reported planning setup.

\begingroup
\setlength{\parskip}{0pt}
\setlength{\IEEEiedtopsep}{2pt}
To address the challenges, we propose CAMP, a cooperative arm--hand motion planning framework for constrained environments. CAMP first generates multiple paths in the lower-dimensional arm space to provide diverse global guides. Along each guide, a layered hand search constructs a complete arm--hand candidate, while local arm relaxation adjusts the guide when neighboring hand configurations cannot be connected collision-free. The resulting candidates are encoded using endpoint-preserving via-point movement primitives (VMPs)~\cite{zhou2019vmp}, yielding a compact representation for coarse-to-fine joint trajectory optimization. Both arm and hand trajectories remain adjustable throughout optimization, allowing CAMP to preserve global route diversity while refining their cooperative motion.

The main contributions of this work are:
\begin{enumerate}
    \item Based on the projection--fiber perspective, we characterize collision-induced arm--hand coupling through configuration-dependent feasible hand fibers, providing a geometric basis for jointly adapting the arm path and hand motion during path lifting.
    \item We propose CAMP, which combines multiple arm-space guides with layered hand search and local arm relaxation, and uses a compact endpoint-preserving VMP representation for coarse-to-fine cooperative trajectory optimization.
    \item We evaluate CAMP across six constrained simulation tasks, achieving $84.2$--$98.5\%$ planning success and outperforming alternative planners with competitive efficiency. Ablation studies verify the key components, and real-robot experiments validate CAMP on constrained manipulation tasks.
\end{enumerate}
\par\endgroup
}

%% file: Sections/3_formulation.tex
\section{Problem Formulation}
\label{sec:problem_formulation}

Consider an arm--hand system consisting of a robot arm and a multi-fingered
hand. The arm and hand have $n_a$ and $n_h$ active joints, respectively, with
configurations $\mathbf q_a\in\mathcal C_a\subseteq\mathbb R^{n_a}$ and
$\mathbf q_h\in\mathcal C_h\subseteq\mathbb R^{n_h}$. The configuration of
the complete system is
\begin{equation}
    \mathbf q
    =
    \begin{bmatrix}
        \mathbf q_a^{\mathsf T} & \mathbf q_h^{\mathsf T}
    \end{bmatrix}^{\mathsf T}
    \in\mathcal C,
    \quad
    \mathcal C=\mathcal C_a\times\mathcal C_h,
    \quad
    n=n_a+n_h.
    \label{eq:joint_configuration_space}
\end{equation}

Let $\mathcal O$ denote the set of workspace obstacles, $B(\mathbf q)$ the
workspace geometry occupied by the complete robot at $\mathbf q$, and
$\chi_{\mathrm{self}}(\mathbf q)$ the indicator of a prohibited
self-collision. The collision-free configuration space of the complete
system is
\begin{equation}
    \mathcal C_{\mathrm{free}}
    =
    \left\{
        \mathbf q\in\mathcal C
        \;\middle|\;
        B(\mathbf q)\cap\mathcal O=\varnothing,
        \quad
        \chi_{\mathrm{self}}(\mathbf q)=0
    \right\}.
    \label{eq:joint_collision_free_space}
\end{equation}

\begin{figure}[!tbp]
    \centering
    \setlength{\abovecaptionskip}{2pt}
    \includegraphics[width=\linewidth]{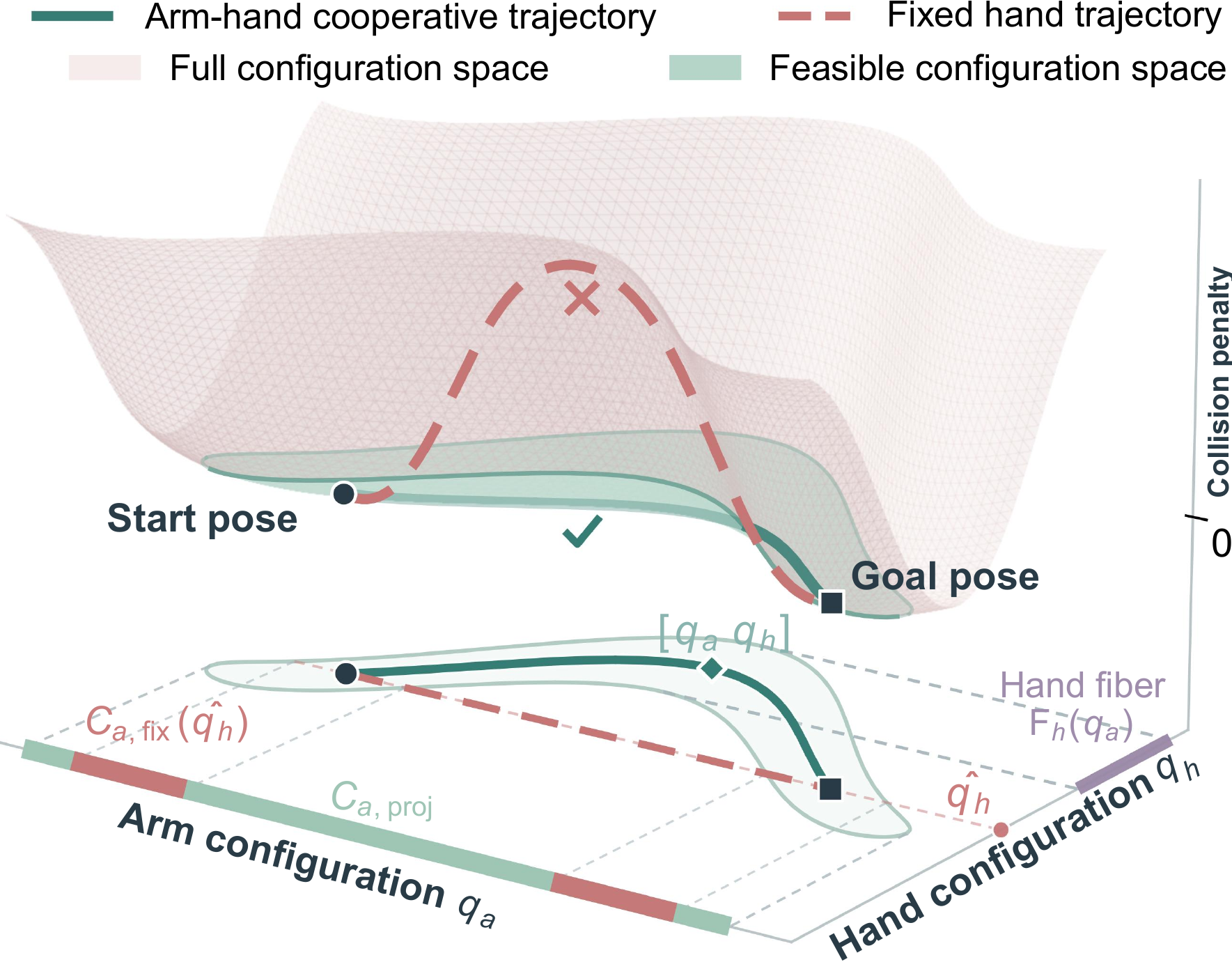}\par
    \nointerlineskip
    \caption{Illustration of the feasible fiber formulation.
    The 3D surface highlights the collision landscape of the \textcolor{feasiblegreen}{feasible configuration space}
    within the \textcolor{spacerose}{full configuration space}.
    The \textcolor{fixedred}{fixed-hand $\hat q_h$ trajectory} leaves the feasible space along part of its path, whereas the
    \textcolor{coopgreen}{arm--hand cooperative trajectory} adapts the hand configuration to remain feasible throughout.
    The bottom projection illustrates how fixing {$\hat q_h$}
    restricts the feasible arm set, while cooperation permits the larger set:
    $\textcolor{fixedred}{\mathcal{C}_{a,\mathrm{fix}}(\hat q_h)}
    \subseteq \textcolor{feasiblegreen}{\mathcal{C}_{a,\mathrm{proj}}}$.}
    \label{fig:arm_hand_coordination}
\end{figure}

Using the classical projection--fiber terminology~\cite{husemoller1994fibre}
and its application to multilevel motion planning~\cite{orthey2024multilevel},
we characterize collision-induced
arm--hand coupling by projecting the joint collision-free space onto the arm
configuration space and defining configuration-dependent feasible hand
fibers. The arm projection $\pi_a:\mathcal C\rightarrow\mathcal C_a$
satisfies $\pi_a(\mathbf q_a,\mathbf q_h)=\mathbf q_a$. For a given arm
configuration $\mathbf q_a$, its feasible hand fiber is
\begin{equation}
    \mathcal F_h(\mathbf q_a)
    =
    \left\{
        \mathbf q_h\in\mathcal C_h
        \;\middle|\;
        \begin{bmatrix}
            \mathbf q_a^{\mathsf T} & \mathbf q_h^{\mathsf T}
        \end{bmatrix}^{\mathsf T}
        \in\mathcal C_{\mathrm{free}}
    \right\}.
    \label{eq:feasible_hand_fiber}
\end{equation}

Although $\mathcal C=\mathcal C_a\times\mathcal C_h$,
$\mathcal C_{\mathrm{free}}$ is generally nonseparable because
$\mathcal F_h(\mathbf q_a)$ varies with $\mathbf q_a$.

The projection of the complete collision-free space onto the arm space and
the arm-feasible set associated with a fixed hand configuration
$\hat{\mathbf q}_h$ are, respectively,
\begin{equation}
    \begin{aligned}
        \mathcal C_{a,\mathrm{proj}}
        &=\pi_a(\mathcal C_{\mathrm{free}})
        =\left\{
            \mathbf q_a\in\mathcal C_a
            \mid
            \mathcal F_h(\mathbf q_a)\neq\varnothing
        \right\}, \\
        \mathcal C_{a,\mathrm{fix}}(\hat{\mathbf q}_h)
        &=\left\{
            \mathbf q_a\in\mathcal C_a
            \mid
            \hat{\mathbf q}_h\in\mathcal F_h(\mathbf q_a)
        \right\}
        \subseteq\mathcal C_{a,\mathrm{proj}}.
    \end{aligned}
    \label{eq:arm_feasible_sets}
\end{equation}

Compared with cooperative planning that allows the hand shape to vary, a fixed hand shape restricts the arm to configurations collision-free with that shape and may therefore reduce the feasible arm set. For a fixed arm path, nonempty feasible hand fibers do not by themselves ensure a continuous collision-free hand motion between the prescribed endpoints. These restrictions explain the coupling underlying the Arm--Hand Planning Dilemma.

Accordingly, let the initial and goal configurations be
$\mathbf q_{\mathrm{init}},\mathbf q_{\mathrm{goal}}\in
\mathcal C_{\mathrm{free}}$. The arm--hand cooperative motion planning problem
over continuous, piecewise differentiable trajectories is
\begin{equation}
    \begin{aligned}
        \text{find}\quad
        &\mathbf q_a(\cdot),\;\mathbf q_h(\cdot) \\
        \text{subject to}\quad
        &\mathbf q(0)=\mathbf q_{\mathrm{init}},
        \qquad
        \mathbf q(1)=\mathbf q_{\mathrm{goal}}, \\
        &\mathbf q_h(\tilde t)\in
        \mathcal F_h\!\left(\mathbf q_a(\tilde t)\right),
        \qquad \forall \tilde t\in[0,1].
    \end{aligned}
    \label{eq:cooperative_planning_problem}
\end{equation}

\begin{figure*}[!t]
    \centering
    \setlength{\abovecaptionskip}{2pt}
    \includegraphics[width=\textwidth]{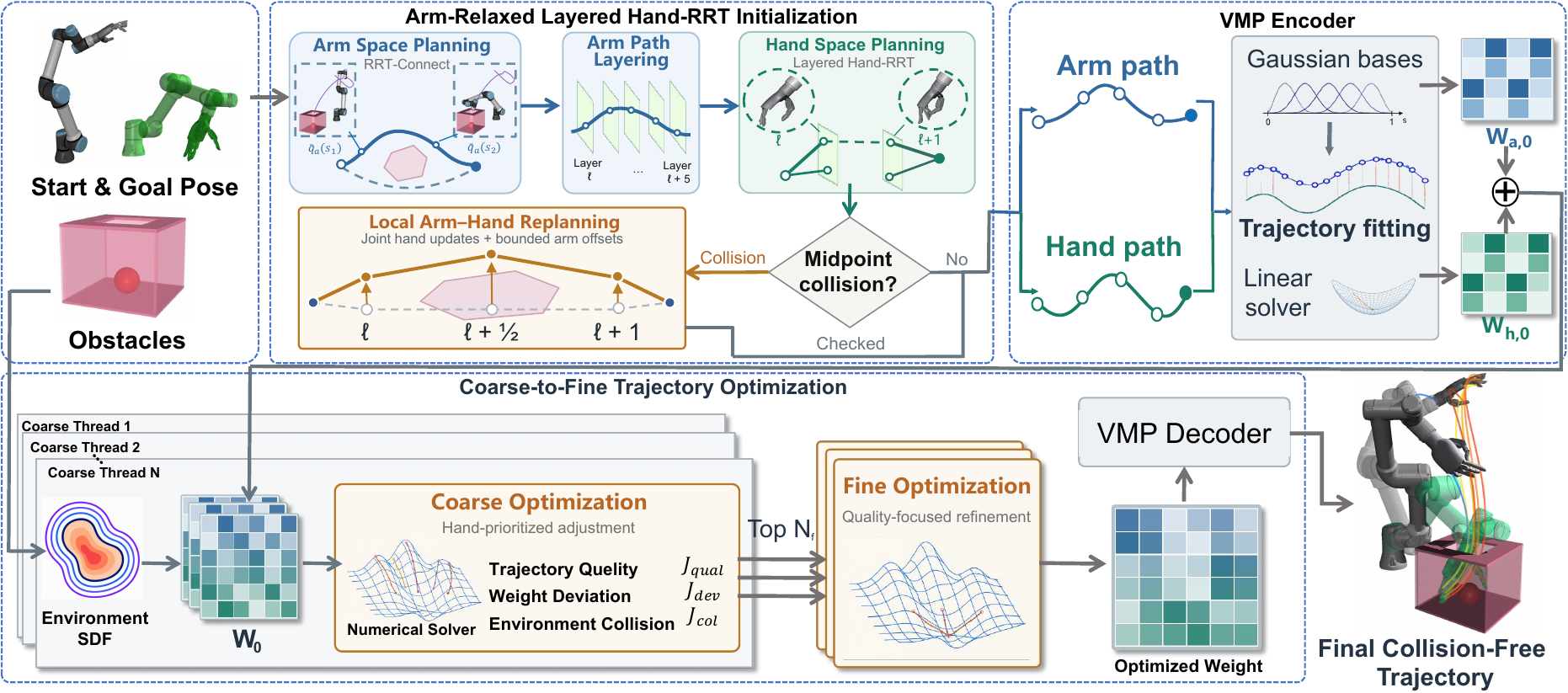}\par
    \nointerlineskip
    \caption{Overview of CAMP. Arm-space guides are lifted to complete
    arm--hand paths through layered Hand-RRT and local joint replanning,
    encoded as VMP weights, and refined by parallel coarse-to-fine
    optimization. The top $N_f$ candidates are further refined before
    the final trajectory is decoded.}
    \label{fig:method_overview}
\end{figure*}

%% file: Sections/4_methodology.tex
\section{Methodology}

This section presents CAMP for the problem defined in Sec.~\ref{sec:problem_formulation}, covering layered hand search with local arm relaxation, VMP trajectory representation, and coarse-to-fine joint optimization of multiple candidates.

\subsection{Arm-Relaxed Layered Hand-RRT Initialization}

RRT-Connect~\cite{kuffner2000rrtconnect} generates a set $\mathcal P_a$ of
candidate reference paths in the arm configuration space between the prescribed start and goal
configurations. We describe initialization for one such path,
$\bar{\mathbf q}_a(\tilde t)$, where $\tilde t\in[0,1]$ denotes normalized progress along
the arm reference path.

The arm reference path provides the overall direction of motion and organizes
the hand search by path progress. We discretize the path parameter on an
ordered grid $0=\tilde t_0<\cdots<\tilde t_{L-1}=1$, with each sample defining a
search layer. Layer $\ell$ uses $\bar{\mathbf q}_a(\tilde t_\ell)$ as its arm
reference configuration. Applying this layering to each path in $\mathcal P_a$
yields the collection of layered paths $\bar{\mathcal P}_a$.

With the arm reference path fixed, a layered bidirectional RRT initializes
two trees at the prescribed initial and goal hand configurations. It samples
hand configurations that satisfy the collision constraints at each layer and
constructs a hand path by connecting adjacent layers under a step-size bound.
The initial search operates on the $n_h$ hand joints.

The arm configurations along the reference path induce the feasible hand
fibers $\mathcal F_h(\bar{\mathbf q}_a(\tilde t))$ defined
in~\eqref{eq:feasible_hand_fiber}. For the fixed arm reference path, the set
of continuous hand paths satisfying the collision constraints is
\begin{equation}
\begin{gathered}
\mathcal L\!\left(\bar{\mathbf q}_a\right)
=
\bigl\{\mathbf q_h\in C^0([0,1],\mathcal C_h)\;\bigm|\\
\mathbf q_h(\tilde t)\in
\mathcal F_h\!\left(\bar{\mathbf q}_a(\tilde t)\right),
\quad \forall \tilde t\in[0,1]\bigr\}.
\end{gathered}
\label{eq:arm_guide_lift}
\end{equation}
Here, $C^0([0,1],\mathcal C_h)$ denotes the space of continuous hand paths.
The paths in this set must connect the prescribed initial and goal hand
configurations.

The hand configuration at each layer lies in its corresponding feasible
fiber, and a feasible transition maintains this membership between layers.
After constructing the hand path, we check the interpolated midpoint
$\tilde t_{\ell+\frac12}$ between layers $\ell$ and $\ell+1$.
The fixed arm reference restricts the feasible hand fibers, which can
prevent a continuous hand connection across a blocked interval.
When the interpolated midpoint is in collision, local arm relaxation
perturbs the arm configuration to explore the feasible fibers induced by nearby
configurations and improve the connection. Specifically, local arm--hand
replanning jointly adjusts the hand configuration $\mathbf q_h(\tilde t)$ and
arm offset $\boldsymbol\delta_a(\tilde t)$ at
$\tilde t_\ell$, $\tilde t_{\ell+\frac12}$, and $\tilde t_{\ell+1}$.

Within this local window, the hand configuration
$\mathbf q_h(\tilde t)\in\mathbb R^{n_h}$ and arm offset
$\boldsymbol\delta_a(\tilde t)\in\mathbb R^{n_a}$ define the complete arm--hand
configuration and its arm offset bound as
\begin{equation}
\begin{aligned}
\bar{\mathbf q}(\tilde t)
&=
\begin{bmatrix}
\bar{\mathbf q}_a(\tilde t)+\boldsymbol\delta_a(\tilde t)\\
\mathbf q_h(\tilde t)
\end{bmatrix}
\in\mathbb R^{n},\\
|\boldsymbol\delta_a(\tilde t)|&\leq\mathbf b_a,
\qquad
\tilde t\in\left\{\tilde t_\ell,\tilde t_{\ell+\frac12},\tilde t_{\ell+1}\right\}.
\end{aligned}
\label{eq:arm_hand_seed_configuration}
\csname ltx@label\endcsname{eq:arm_relaxed_hand_rrt_constraints}
\end{equation}
The componentwise interval $[-\mathbf b_a,\mathbf b_a]$ bounds each arm
joint offset, keeping the arm configurations at the midpoint and its two
neighboring layers close to their respective references.
Local replanning uses the layered bidirectional RRT to jointly sample
$\mathbf q_h(\tilde t)$ and $\boldsymbol\delta_a(\tilde t)$ at
$\tilde t_\ell$, $\tilde t_{\ell+\frac12}$, and $\tilde t_{\ell+1}$.
Original path nodes outside the local window serve as connection anchors,
and the search checks both the local path and its connections to the
original path.

Splicing the locally replanned segments back into the original path yields
the complete arm--hand trajectory:
\begin{equation}
\bar{\boldsymbol{\tau}}(\tilde t)=
\begin{bmatrix}
\bar{\mathbf q}_a(\tilde t)+\boldsymbol{\delta}_a(\tilde t)\\
\mathbf q_h(\tilde t)
\end{bmatrix},
\qquad \tilde t\in[0,1].
\label{eq:arm_hand_seed_trajectory}
\end{equation}

\begin{algorithm}[!tbp]
\caption{Arm--Hand Cooperative Motion Planning}
\label{alg:cooperative_motion_planning}
\DontPrintSemicolon
\KwIn{System $\mathcal R$, obstacles $\mathcal O$,
start configuration $\mathbf q_{\mathrm{init}}$,
and goal configuration $\mathbf q_{\mathrm{goal}}$.}
\KwOut{Planned arm--hand trajectory $\boldsymbol\tau^*$.}

$\mathcal P_a \leftarrow
\operatorname{RRT\text{-}Connect}
(\mathbf q_{a,\mathrm{init}},
 \mathbf q_{a,\mathrm{goal}},\mathcal O)$\;

$\bar{\mathcal P}_a \leftarrow
\operatorname{ArmPathLayering}(\mathcal P_a)$\;

$\bar{\mathcal T}_0 \leftarrow
\operatorname{LayeredHand\text{-}RRT}
(\bar{\mathcal P}_a,\mathbf q_{h,\mathrm{init}},
 \mathbf q_{h,\mathrm{goal}})$\;

$\mathcal I \leftarrow
\operatorname{CheckMidpoints}(\bar{\mathcal T}_0)$\;

\If{$\mathcal I\neq\varnothing$}{
    $\bar{\mathcal T}_0 \leftarrow
    \operatorname{LocalArmHandReplanning}
    (\bar{\mathcal T}_0,\mathcal I,\mathbf b_a)$\;
}

$\mathcal W_0 \leftarrow
\operatorname{EncodeVMP}(\bar{\mathcal T}_0)$\;

$\mathcal W_c \leftarrow
\operatorname{CoarseOptimizeParallel}(\mathcal W_0,J_c,K_c)$\;

$\mathcal W_c^* \leftarrow
\operatorname{RankAndSelect}(\mathcal W_c,N_f)$\;

$\mathcal W_f \leftarrow
\operatorname{FineOptimizeParallel}(\mathcal W_c^*,J_f,K_f)$\;

$\mathcal T_f \leftarrow
\operatorname{DecodeVMP}(\mathcal W_f)$\;

$\boldsymbol\tau^* \leftarrow
\operatorname{SelectBest}(\mathcal T_f)$\;

\Return{$\boldsymbol\tau^*$}\;
\end{algorithm}

\subsection{VMP Representation}

CAMP represents the complete arm--hand trajectory with VMP~\cite{zhou2019vmp} phase basis
functions and weights. For a given complete candidate generated by the
initialization stage, the arm and hand trajectories are decoded as
\begin{equation}
\begin{aligned}
\mathbf{q}_a(\tilde t)
&=
\mathbf{h}_a(\tilde t)
+
g(\tilde t)\mathbf{W}_a^{\mathsf T}\boldsymbol{\phi}_a(\tilde t), \\
\mathbf{q}_h(\tilde t)
&=
\mathbf{h}_h(\tilde t)
+
g(\tilde t)\mathbf{W}_h^{\mathsf T}\boldsymbol{\phi}_h(\tilde t),
\qquad \tilde t\in[0,1].
\end{aligned}
\label{eq:arm_hand_vmp_representation}
\end{equation}
For subsystem $r\in\{a,h\}$, $\mathbf{h}_r(\tilde t)$ is the linear endpoint
reference, $\boldsymbol{\phi}_r(\tilde t)\in\mathbb{R}^{K_r}$ is the normalized
Gaussian basis vector, and $g(\tilde t)$ is an endpoint envelope that vanishes at the
start and goal so that the deformation term preserves both endpoints. The
weights $\mathbf{W}_a$ and
$\mathbf{W}_h$ control the arm trajectory
deformation and hand-shape evolution, respectively.

Smooth basis functions restrict trajectory deformation to a low-dimensional
temporal subspace, thereby suppressing the high-frequency local changes allowed
by waypoint-wise parameterization. With the first and last waypoints fixed, the
number of scalar decision variables in the waypoint-wise and VMP
parameterizations are
\begin{equation}
\begin{aligned}
    N_{\mathrm{joint}}
    &=
    (T-2)(n_a+n_h), \\
    N_{\mathrm{VMP}}
    &=
    K_a n_a+K_h n_h .
\end{aligned}
\label{eq:trajectory_parameter_counts}
\end{equation}

Here, $K_a$ and $K_h$ denote the numbers of arm and hand basis functions. Compared with independent waypoint updates, VMP weights couple neighboring waypoint deformations through smooth basis functions and reduce the optimization dimension when $N_{\mathrm{VMP}}\ll N_{\mathrm{joint}}$. Separate choices of $K_a$ and $K_h$ control the relative sizes of the arm and hand weight spaces: $K_a>K_h$ provides finer temporal detail for local arm obstacle avoidance and a more compact basis for overall hand-shape changes.

VMP initialization encodes the full arm--hand candidate trajectories from Section~A through regularized fitting after endpoint-preserving resampling on the common grid $\{\tilde t_t\}_{t=0}^{T-1}$. For each subsystem $r\in\{a,h\}$, let $\bar{\mathbf Q}_r$ denote the resampled trajectory matrix, $\mathbf H_r$ the linear reference matrix connecting the endpoints, and $\mathbf B_r$ the basis matrix whose $t$th row is $g(\tilde t_t)\boldsymbol\phi_r(\tilde t_t)^{\mathsf T}$. The initial weights are obtained from the same fitting objective:
\begin{equation}
\begin{aligned}
\mathbf{W}_{r,0}
&=
\underset{\mathbf{W}_r}{\arg\min}\;
\left\|
\mathbf{H}_r+\mathbf{B}_r\mathbf{W}_r
-\bar{\mathbf{Q}}_r
\right\|_F^2
\\
&\quad+
\lambda_{\mathrm{w}}
\left\|\mathbf{W}_r\right\|_F^2
\\
&\quad+
\lambda_{\mathrm{c}}
\left\|
\mathbf{D}
\left(\mathbf{H}_r+\mathbf{B}_r\mathbf{W}_r\right)
\right\|_F^2 .
\end{aligned}
\label{eq:vmp_weight_encoding_objective}
\end{equation}

The three terms penalize trajectory reconstruction error, weight magnitude, and second-order variation of the reconstructed trajectory, respectively. Here, $\mathbf D$ is the second-order difference matrix, and $\lambda_{\mathrm w}>0$ and $\lambda_{\mathrm c}\geq0$ control weight and smoothness regularization, respectively. This regularized least-squares problem is quadratic in the VMP weights and can be solved directly through a linear system. The resulting weights form the initialization $\mathbf W_0=(\mathbf W_{a,0},\mathbf W_{h,0})$.

\begingroup
\setlength{\parskip}{0pt plus 0.3pt}
\setlength{\abovedisplayskip}{1.5ex plus 1pt}
\setlength{\belowdisplayskip}{1.5ex plus 1pt}
\setlength{\abovedisplayshortskip}{0pt plus 1pt}
\setlength{\belowdisplayshortskip}{1.5ex plus 1pt}
\makeatletter
\renewcommand{\subsection}{\@startsection{subsection}{2}{\z@}{1.5ex}{0.7ex}{\normalfont\normalsize\itshape}}
\makeatother
\subsection{Arm--Hand Joint Trajectory Optimization}

The coarse and fine stages optimize the joint weights
$\mathbf{W}=(\mathbf{W}_a,\mathbf{W}_h)$ using the same objective terms
with stage-specific weights. With $\rho\in\{c,f\}$ denoting the coarse
and fine stages, respectively, the objective is
\begin{equation}
\begin{aligned}
    J_{\rho}\!\left(\mathbf{W}\right)
    ={}& \lambda_{\mathrm{qual}}^{\rho}
    J_{\mathrm{qual}}\!\left(\mathbf{W}\right)
    + J_{\mathrm{dev},\rho}\!\left(\mathbf{W}\right) \\
    &+ \lambda_{\mathrm{col}}^{\rho}
    J_{\mathrm{col}}\!\left(\mathbf{W}\right).
\end{aligned}
\label{eq:joint_weight_objective}
\end{equation}
\begin{equation}
J_{\mathrm{qual}}\!\left(\mathbf{W}\right)
=
\left\|
\mathcal{D}_{\mathrm{qual}}\!\left[
\mathbf{q}_{0:T-1}
\right]
\right\|_F^2 .
\label{eq:trajectory_quality_cost}
\end{equation}
\begin{equation}
\begin{aligned}
J_{\mathrm{dev},\rho}\!\left(\mathbf{W}\right)
={}&
\lambda_{\mathrm{dev},a}^{\rho}
\left\|
\mathbf{w}_a-\mathbf{w}_{a,0}
\right\|_2^2 \\
&+
\lambda_{\mathrm{dev},h}^{\rho}
\left\|
\mathbf{w}_h-\mathbf{w}_{h,0}
\right\|_2^2 .
\end{aligned}
\label{eq:weight_deviation_cost}
\end{equation}
\begin{equation}
J_{\mathrm{col}}\!\left(\mathbf{W}\right)
=
\left\|
\mathbf{r}_{\mathrm{env}}^{\mathrm{soft}}\!\left(
\mathbf{q}_{0:T-1}
\right)
\right\|_2^2 .
\label{eq:soft_collision_cost}
\end{equation}
Here, $\mathcal{D}_{\mathrm{qual}}$ is a weighted trajectory-quality operator
that combines adjacent configuration differences and five-point
finite-difference accelerations to penalize excessive joint variation and
high-frequency oscillations. The vectors $\mathbf{w}_a$ and
$\mathbf{w}_h$ are the vectorized arm and hand weights. The joint
deviation cost softly anchors them to $\mathbf{w}_{a,0}$ and
$\mathbf{w}_{h,0}$, with stage-dependent coefficients controlling their
relative influence. Both weight blocks therefore remain adjustable in both
stages. The soft environment-collision term penalizes trajectory states near
or inside obstacles and provides an optimization direction away from the
environment.

\begin{figure}[!tbp]
    \centering
    \setlength{\abovecaptionskip}{2pt}
    \includegraphics[width=\columnwidth]{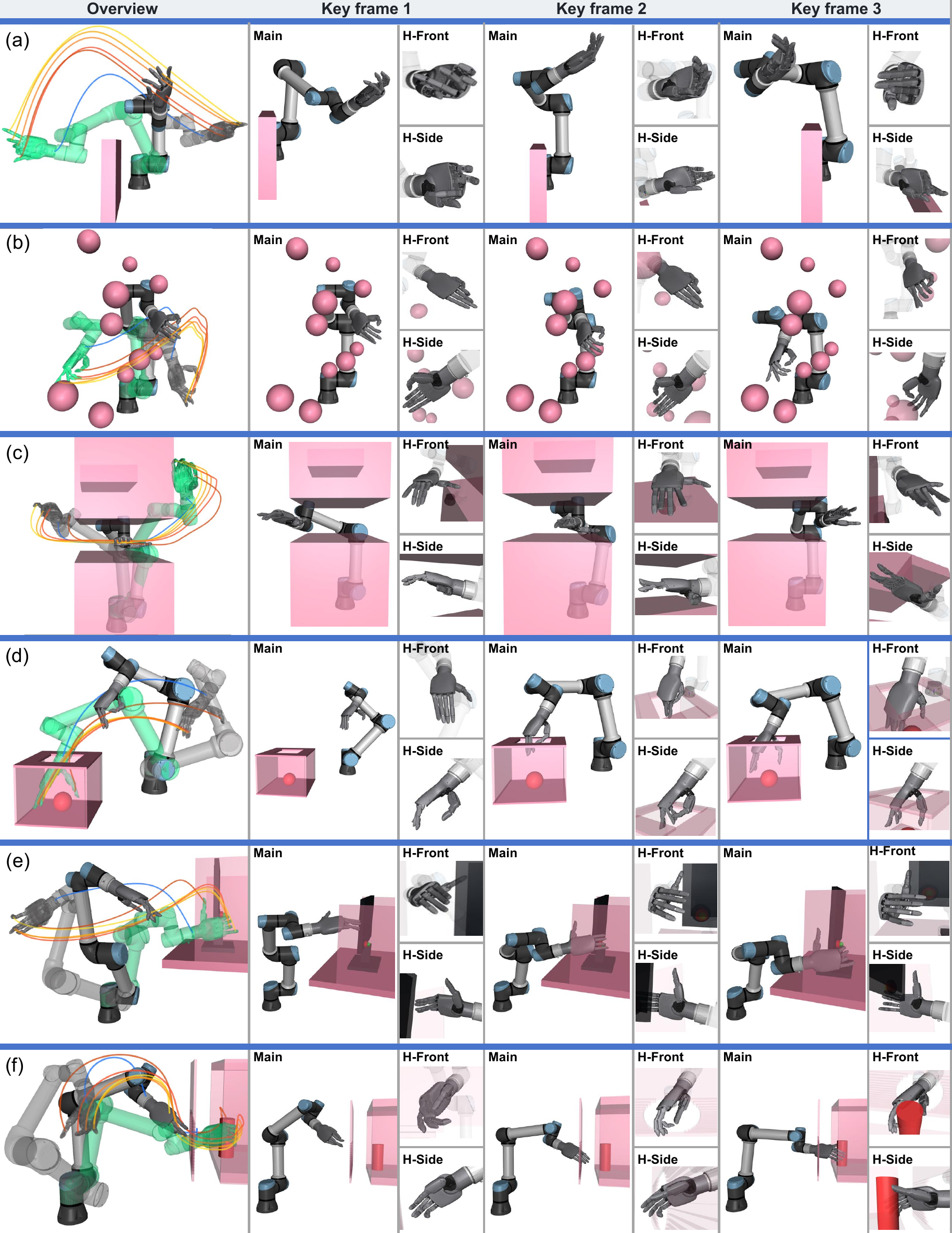}\par
    \nointerlineskip
    \caption{Six simulated scenarios: (a) Wall Traversal, (b) Multi-Sphere Avoidance, (c) Narrow Passage Traversal, (d) Ball-in-Box Pregrasp, (e) Display Button Press, and (f) Cabinet Cylinder Pregrasp. The \textcolor[HTML]{8F8D8D}{gray} and \textcolor[HTML]{21FD96}{green} robot models denote the start and goal configurations, respectively. \textcolor[HTML]{1E90FF}{C}\textcolor[HTML]{F4511E}{o}\textcolor[HTML]{FF5A36}{l}\textcolor[HTML]{FF922B}{o}\textcolor[HTML]{FFB300}{r}\textcolor[HTML]{FFD633}{e}\textcolor[HTML]{FFD633}{d} dotted traces show the trajectories of the wrist and fingertips, and \textcolor[HTML]{FD94B2}{pink} geometry denotes obstacles.}
    \label{fig:simulation_environments}
\end{figure}

Under the feasible-fiber formulation, the environment- and self-collision
constraints require the decoded hand configuration at every discrete trajectory
node to lie in the feasible hand fiber induced by the decoded arm configuration:
\begin{equation}
\begin{gathered}
\mathbf{q}_h\!\left(\tilde t_t;\mathbf{W}_h\right)
\in
\mathcal F_h\!\left(
\mathbf{q}_a\!\left(\tilde t_t;\mathbf{W}_a\right)
\right),\\
t=0,\ldots,T-1.
\end{gathered}
\label{eq:discrete_fiber_membership}
\end{equation}
Equation~\eqref{eq:discrete_fiber_membership} induces a phase-indexed sequence
of feasible hand fibers along the decoded arm trajectory. The arm weights
$\mathbf{W}_a$ determine this sequence by changing
$\mathbf{q}_a(\tilde t_t)$, whereas the hand weights
$\mathbf{W}_h$ determine the hand configuration selected at each node.
Because $\mathcal F_h(\mathbf q_a)$ varies with the arm configuration, changing
the arm weights also changes the feasible set available to the hand weights.
The two variable blocks are therefore inseparable at the constraint level.
Fixing $\mathbf{W}_a$ reduces the problem to finding a hand path through
a prescribed fiber sequence. Jointly optimizing $\mathbf{W}_a$ and
$\mathbf{W}_h$ allows both the fiber sequence and the hand configurations
selected within it to change, which constitutes the core arm--hand coupling.

Let $\mathbf q_t=\mathbf q(\tilde t_t;\mathbf W)$. The cooperative
trajectory optimization for the given candidate at stage $\rho$ is
\begin{equation}
\begin{aligned}
    \mathbf{W}_{\rho}^{*}
    ={}&
    \operatorname*{arg\,min}_{\mathbf{W}}
    \; J_{\rho}\!\left(\mathbf{W}\right) \\
    \mathrm{s.t.}\quad
    & \mathbf{q}_{\min}
      \leq \mathbf{q}_t
      \leq \mathbf{q}_{\max}, \\
    & \left|
        \frac{
            \mathbf{q}_{t+1}-\mathbf{q}_{t}
        }{\Delta t_t}
      \right|
      \leq \dot{\mathbf{q}}_{\max}, \\
    & \mathbf{r}_{\mathrm{env}}(\mathbf{q}_t)\leq\mathbf{0},
      \qquad
      \mathbf{r}_{\mathrm{self}}(\mathbf{q}_t)\leq\mathbf{0}.
\end{aligned}
\label{eq:joint_weight_optimization}
\end{equation}
Here, $\mathbf{q}_{\min}$ and $\mathbf{q}_{\max}$ are the joint lower and upper
limits of the complete system, $\dot{\mathbf{q}}_{\max}$ is the joint velocity
limit, and $\Delta t_t$ is the time interval between adjacent phase nodes.
The functions $\mathbf{r}_{\mathrm{env}}$ and $\mathbf{r}_{\mathrm{self}}$
denote the environment- and self-collision residuals, respectively.
\par\endgroup

\subsection{Coarse-to-Fine Multi-Candidate Lift Optimization}

CAMP refines the complete arm--hand candidates generated by the
initialization stage through coarse-to-fine optimization. These candidates
capture different arm routes and their associated hand motions.
Let $\mathcal{W}_0$ denote the collection of initialized candidates,
represented by their current weights. Each candidate retains its own
reference trajectories and initial weights throughout optimization,
selection, and decoding.

Let $\mathcal{S}_{\rho,K}(\mathbf{W})$ denote the weights returned for a
given candidate after at most $K$ iterations from $\mathbf{W}$, using
its associated references, initial weights, the stage objective $J_{\rho}$,
and the common hard constraints defined in Sec.~IV-C.
The coarse stage jointly optimizes the VMP-encoded arm--hand candidates,
prioritizing the reduction of collision and other constraint violations.
It uses the smaller iteration budget $K_c$ to assess candidate feasibility.
A stronger penalty on deviations from the initial arm weights favors
hand-dominant corrections while allowing necessary arm adjustments.
Parallel coarse optimization of all candidates is written as:
\begin{equation}
\mathcal{W}_c
=
\left\{
\mathcal{S}_{c,K_c}\!\left(\mathbf{W}_0\right)
\,\middle|\,
\mathbf{W}_0\in\mathcal{W}_0
\right\}.
\label{eq:parallel_coarse_weight_optimization}
\end{equation}

After coarse optimization, numerically invalid candidates are discarded, and
the remaining candidates are evaluated on the same discrete phase grid.
Candidates satisfying all joint-limit, velocity-limit, environment-collision,
and self-collision constraints receive priority. The candidates are then
ranked by constraint satisfaction and trajectory quality, and the top $N_f$
coarse candidates form the collection $\mathcal{W}_{c}^{*}$.

The fine stage starts from the $N_f$ selected coarse candidates, increases
the relative weight of trajectory quality, and reduces the soft penalty on
deviations from the initial arm weights. It jointly adjusts the arm and hand
weights with the larger iteration budget $K_f$ to further improve feasibility
and trajectory quality:
\begin{equation}
\mathcal{W}_{f}
=
\left\{
\mathcal{S}_{f,K_f}\!\left(\mathbf{W}_{c}\right)
\;\middle|\;
\mathbf{W}_{c}\in\mathcal{W}_{c}^{*}
\right\},
\qquad K_c<K_f .
\label{eq:fine_weight_optimization}
\end{equation}
The candidates in $\mathcal{W}_{f}$ are decoded using their associated references, and
the trajectory with the best quality is returned as $\boldsymbol{\tau}^{*}$.
The value of $N_f$ is specified in the experimental setup.

%% file: Sections/5_experiment_setup.tex
\section{Experiments}

 We design the experiments to answer the following questions: 1) Does arm-hand cooperative planning improve feasibility over decoupled planning? 2) How does CAMP compare with alternative motion planners? 3) How do local arm relaxation, VMP representation, and coarse-to-fine optimization affect planning performance? 4) Can the resulting trajectories be executed on the real robot system?

\textbf{Experimental Setup:} The platform comprises a 6-DoF UR7e arm and a 16-DoF LinkerHand. All methods run on a workstation with an Intel Core i9-14900KF CPU and an NVIDIA GeForce RTX 5070 Ti GPU. During planning, PyRoKi~\cite{kim_pyroki_2025} computes the environment's signed distance field and approximates robot links as capsules. We validate final trajectories using exact collision checks in MuJoCo~\cite{todorov2012mujoco}. QRRT*~\cite{orthey2024multilevel} and RRT-Connect~\cite{kuffner2000rrtconnect} use OMPL~\cite{guo2026the-open-motion-planning-library2}.

The six simulated tasks in Fig.~\ref{fig:simulation_environments} evaluate motion planning to prescribed goal configurations. In each trial, we sample obstacle configurations from predefined task-specific ranges.We specify target positions for multiple fingertips and use IK-Beam~\cite{kim_pyroki_2025} to compute candidate goal configurations, which we manually screen and adjust for feasibility. All methods use identical scenes and start and goal configurations.

\textbf{Implementation Details:} Each task--method batch contains 100 trials. Trajectories use 200 waypoints, with 30 VMP basis functions per arm joint and 20 per hand joint. Initialization runs 16 parallel instances with $L=20$ layers, a 10.0~s limit per RRT search, and arm-joint offsets within $[-0.1,0.1]$~rad of their references. We fit VMP weights for up to eight candidates before parallel coarse optimization, then refine the top $N_f=3$. The coarse and fine iteration limits are $K_c=100$ and $K_f=200$, respectively.

\begin{table}[t]
\centering
\footnotesize
\renewcommand{\arraystretch}{1.05}
\setlength{\tabcolsep}{3pt}
\caption{Necessity of arm-hand cooperative planning.}
\label{tab:necessity_cooperation}
\begin{tabular}{@{}ccc@{}}
\toprule
\textbf{Task} & \textbf{Method} & \textbf{S.R.} (\%) $\uparrow$ \\
\midrule
\multirow{3}{*}{\shortstack{Narrow Passage\\Traversal}}
 & $A\!\rightarrow\!H$                 & 37 \\
 & $P\!\rightarrow\!A\!\rightarrow\!H$ & 85 \\
 & \cellcolor{black!6}\textbf{CAMP}      & \cellcolor{black!6}\textbf{93} \\
\addlinespace[2pt]
\multirow{3}{*}{\shortstack{Ball-in-Box\\Pregrasp}}
 & $A\!\rightarrow\!H$                 & 4  \\
 & $P\!\rightarrow\!A\!\rightarrow\!H$ & 89 \\
 & \cellcolor{black!6}\textbf{CAMP}      & \cellcolor{black!6}\textbf{92} \\
\addlinespace[2pt]
\multirow{3}{*}{\shortstack{Display Button\\Press}}
 & $A\!\rightarrow\!H$                 & 16 \\
 & $P\!\rightarrow\!A\!\rightarrow\!H$ & 59 \\
 & \cellcolor{black!6}\textbf{CAMP}      & \cellcolor{black!6}\textbf{94} \\
\addlinespace[2pt]
\multirow{3}{*}{\shortstack{Cabinet Cylinder\\Pregrasp}}
 & $A\!\rightarrow\!H$                 & 1  \\
 & $P\!\rightarrow\!A\!\rightarrow\!H$ & 6  \\
 & \cellcolor{black!6}\textbf{CAMP}      & \cellcolor{black!6}\textbf{84} \\
\bottomrule
\end{tabular}
\par
\end{table}

\begin{table*}[t]
\centering
\footnotesize
\setlength{\tabcolsep}{12pt}
\renewcommand{\arraystretch}{1.08}
\caption{Comparison of motion planners over 10 batches of 100 trials per task--method pair. S.R. and P.T. are reported as mean $\pm$ standard deviation across batches. CAMP rows are shaded; best entries are \textbf{bolded} and second-best entries are \underline{underlined}.}
\label{tab:motion_planning_comparison}
\vspace{-0.45\baselineskip}
\begin{tabular}{ccccccc}
\toprule
\multirow{2}{*}{\textbf{Method}} &
\multicolumn{2}{c}{\textbf{Wall Traversal}} &
\multicolumn{2}{c}{\textbf{Multi-Sphere Avoidance}} &
\multicolumn{2}{c}{\textbf{Narrow Passage Traversal}} \\
\cmidrule(lr){2-3}\cmidrule(lr){4-5}\cmidrule(lr){6-7}
& \textbf{S.R. (\%)} $\uparrow$ & \textbf{P.T. (s)} $\downarrow$
& \textbf{S.R. (\%)} $\uparrow$ & \textbf{P.T. (s)} $\downarrow$
& \textbf{S.R. (\%)} $\uparrow$ & \textbf{P.T. (s)} $\downarrow$ \\
\midrule
RRT-Connect    & 94.00 $\pm$ 2.00 & 2.41 $\pm$ 0.28             & 66.30 $\pm$ 5.57 & 8.09 $\pm$ 0.42          & 3.00 $\pm$ 2.24  & 36.82 $\pm$ 0.63             \\
QRRT*          & 96.70 $\pm$ 2.06 & 4.96 $\pm$ 0.70            & 70.40 $\pm$ 2.91 & 16.91 $\pm$ 1.20         & 14.60 $\pm$ 2.99 & 39.83 $\pm$ 1.61          \\
CHOMP          & 63.00 $\pm$ 1.83 & 13.50 $\pm$ 1.47            & 72.00 $\pm$ 1.41 & 22.60 $\pm$ 3.68         & 38.20 $\pm$ 2.06 & 41.60 $\pm$ 3.88          \\
A*+CHOMP       & 84.50 $\pm$ 3.03 & 13.40 $\pm$ 1.36            & 71.50 $\pm$ 3.70 & 22.30 $\pm$ 3.65         & 71.50 $\pm$ 3.70 & 53.80 $\pm$ 5.32          \\
\rowcolor{black!6}
\textbf{CAMP} &
\textbf{98.50 $\pm$ 1.33} &
\underline{3.40 $\pm$ 1.13} &
\textbf{91.30 $\pm$ 1.51} &
\textbf{4.70 $\pm$ 1.47} &
\textbf{92.50 $\pm$ 2.11} &
\textbf{30.10 $\pm$ 1.20} \\
\midrule
\multirow{2}{*}{\textbf{Method}} &
\multicolumn{2}{c}{\textbf{Ball-in-Box Pregrasp}} &
\multicolumn{2}{c}{\textbf{Display Button Press}} &
\multicolumn{2}{c}{\textbf{Cabinet Cylinder Pregrasp}} \\
\cmidrule(lr){2-3}\cmidrule(lr){4-5}\cmidrule(lr){6-7}
& \textbf{S.R. (\%)} $\uparrow$ & \textbf{P.T. (s)} $\downarrow$
& \textbf{S.R. (\%)} $\uparrow$ & \textbf{P.T. (s)} $\downarrow$
& \textbf{S.R. (\%)} $\uparrow$ & \textbf{P.T. (s)} $\downarrow$ \\
\midrule
RRT-Connect    & 0.00 $\pm$ 0.00  & --              & 1.50 $\pm$ 1.20  & 49.37 $\pm$ 0.18          & 0.00 $\pm$ 0.00  & --              \\
QRRT*          & 12.10 $\pm$ 1.29 & 49.26 $\pm$ 0.59            & 23.00 $\pm$ 1.06 & 48.08 $\pm$ 0.60          & 8.10 $\pm$ 1.50 & 49.10 $\pm$ 0.57          \\
CHOMP          & 39.50 $\pm$ 1.00 & 73.60 $\pm$ 5.17           & 23.50 $\pm$ 0.58 & 83.70 $\pm$ 1.11          & 18.30 $\pm$ 2.87 & 79.90 $\pm$ 3.63           \\
A*+CHOMP       & 50.40 $\pm$ 0.50 & 60.55 $\pm$ 18.91           & 45.70 $\pm$ 1.71 & 79.23 $\pm$ 13.77          & 18.80 $\pm$ 2.06 & 98.40 $\pm$ 8.43           \\
\rowcolor{black!6}
\textbf{CAMP} &
\textbf{92.00 $\pm$ 2.09} &
\textbf{32.40 $\pm$ 1.20} &
\textbf{94.40 $\pm$ 2.49} &
\textbf{35.60 $\pm$ 1.47} &
\textbf{84.20 $\pm$ 1.30} &
\textbf{42.60 $\pm$ 2.08} \\
\bottomrule
\end{tabular}
\par
\end{table*}

\begin{table}[t]
\centering
\footnotesize
\setlength{\tabcolsep}{2pt}
\renewcommand{\arraystretch}{1.10}
\caption{Ablation Study}
\label{tab:vmp_ablation}
\vspace{-0.45\baselineskip}
\begin{tabular}{@{}cccccc@{}}
\toprule
\multicolumn{3}{c}{\textbf{Method Configuration}} &
\multicolumn{3}{c}{\textbf{Performance Metrics}} \\
\cmidrule(lr){1-3}\cmidrule(lr){4-6}
\begin{tabular}[c]{@{}c@{}}\textbf{Arm}\\\textbf{Relaxation}\end{tabular} &
\textbf{Representation} &
\textbf{Optimization} &
\textbf{S.R.} (\%) & \textbf{P.T.} (s) & \textbf{N.T.L.} \\
\midrule
Fixed-arm   & VMP   & Coarse2fine  & 74 & 32.68 & 1.614 \\
Arm-relaxed & Joint & Coarse2fine  & 79 & 84.97 & 2.260 \\
Arm-relaxed & VMP   & Fine-only & 81 & 25.15 & 1.679 \\
\rowcolor{black!6}
\textbf{Arm-relaxed} & \textbf{VMP} & \textbf{Coarse2fine} &
\textbf{92} & \underline{32.45} & \underline{1.598} \\
\bottomrule
\end{tabular}
\par
\end{table}

\textbf{Evaluation Metrics:} The following three metrics jointly evaluate planning feasibility, computational efficiency, and trajectory quality. 

\textit{Success Rate (S.R.)} is the percentage of trials in each batch that produce a complete 22-DoF trajectory satisfying the task constraints, reaching the prescribed goal configuration, and passing exact collision checks in MuJoCo.

\textit{Planning Time (P.T.)} is the wall-clock time spent by each method's planning solver. It excludes planner setup, path simplification, post-planning dense validation, collision-checker construction, artifact I/O, and MuJoCo playback. Each batch reports the arithmetic mean over all 100 trials.

\textit{Normalized Joint-Space Trajectory Length (N.T.L.)} measures the cumulative joint variation relative to the direct start--goal displacement:
\begin{equation}
    L_{q}^{\mathrm{norm}} =
    \frac{
        \sum_{t=0}^{T-2}
        \left\|\mathbf{q}_{t+1}-\mathbf{q}_{t}\right\|_{2}
    }{
        \left\|\mathbf{q}_{T-1}-\mathbf{q}_{0}\right\|_{2}
    },
\end{equation}
A value of one corresponds to straight-line interpolation in joint space, while larger values indicate additional joint variation beyond the direct start--goal displacement. N.T.L. is evaluated over the successful trajectories of each method and reported as the arithmetic mean. 

%% file: Sections/6_Experiments.tex
\subsection{Does Arm-Hand Cooperative Planning Improve Feasibility over Decoupled Planning?}

We compare CAMP with two decoupled baselines under the same task setup and evaluation protocol. Arm-Then-Hand ($A\rightarrow H$) plans the arm with the initial hand configuration fixed, then linearly interpolates the hand to the goal. Preshape-Then-Arm-Then-Hand ($P\rightarrow A \rightarrow H$) instead uses a task-dependent preshape during arm planning: an open hand for Narrow Passage Traversal and Display Button Press, and a compact hand for Ball-in-Box Pregrasp and Cabinet Cylinder Pregrasp.

Table~\ref{tab:necessity_cooperation} shows that CAMP achieves the highest
S.R. across all four tasks, which highlights the limitations of
decoupled planning. Fixing the initial hand configuration in
$A\!\rightarrow\!H$ prevents adaptation to spatial constraints that vary along
the path. The task-dependent preshape in
$P\!\rightarrow\!A\!\rightarrow\!H$ provides a more suitable configuration for
some passages, but a single fixed shape cannot accommodate motions that require hand-shape transitions. For example, in Cabinet Cylinder Pregrasp, the hand must remain compact while passing the front barrier and reopen near the cylinder. CAMP coordinates hand reopening with arm motion, allowing the arm pose to adjust for collision avoidance during the transition.

\subsection{How Does CAMP Compare with Alternative Motion Planners?}

We compare CAMP with sampling-based, optimization-based, and hybrid search--optimization baselines. RRT-Connect~\cite{kuffner2000rrtconnect} plans directly in the full 22-DoF arm--hand space. QRRT* uses Section Patterns~\cite{orthey2021sectionpatterns} to lift paths through the three-level finger-removal hierarchy in~\cite{orthey2024multilevel}: arm with the index finger, arm with the index finger and thumb, and the complete arm--hand system. CHOMP~\cite{zucker2013chomp} directly optimize full 22-DoF trajectories, differing only in initialization. CHOMP uses linear joint-space interpolation for initalization. Following the initialization strategy of Jiao et al.~\cite{jiao_consolidating_2021}, A*+CHOMP maps collision-free wrist-position waypoints from A*~\cite{hart_astar_1968} to arm-joint anchors through waypoint IK and combines them with linearly interpolated hand joints to form a full-joint initialization. All methods process candidates in parallel for a fair comparison.

Table~\ref{tab:motion_planning_comparison} shows that CAMP consistently achieves the highest S.R. RRT-Connect degrades in constrained scenes, which is consistent with the difficulty of sampling directly in the complete configuration space. The main reason for QRRT* failure is search timeout and collision in constrained space. Its low absolute S.R. in the more difficult tasks is consistent with the challenge of lifting paths through successive arm--hand bundle levels under strong arm--hand coupling. CAMP addresses this limitation by organizing bidirectional hand search over
ordered layers of multiple arm reference paths and locally relaxing the arm
configuration at blocked inter-layer transitions. The resulting complete
arm--hand candidates are further refined through joint VMP optimization,
which is consistent with the higher S.R. of CAMP in the constrained tasks.

CAMP addresses this difficulty by combining local path lifting with subsequent trajectory-wide joint refinement. Near obstacles accommodating hand motion require coordinated changes to the arm trajectory. Layered bidirectional hand search and local arm relaxation construct complete candidates along multiple arm guides. Joint VMP optimization then coordinates arm clearance and hand posture across neighboring segments, while parallel coarse optimization and selective fine refinement concentrate the larger optimization budget on promising candidates. 

A* initialization improves CHOMP in several constrained
tasks, indicating that global path guidance is important for avoiding poor
local minima. Nevertheless, both optimization baselines rely on a single
full-joint initialization, which limits the route diversity available to the
subsequent local optimization. In contrast, CAMP accounts for arm--hand collision coupling during initialization through layered hand search and local arm relaxation, providing coordinated candidates for subsequent joint VMP optimization.

CAMP also achieves competitive planning efficiency. A key source of this efficiency is the compact VMP representation, which reduces the optimization dimension relative to waypoint-wise joint optimization. The representation ablation in Table~\ref{tab:vmp_ablation} further supports this explanation. CAMP maintains a high S.R. while reducing P.T. in constrained environments, indicating that reduced-dimensional global guidance and compact coarse-to-fine joint optimization provide an effective balance between global exploration and local trajectory refinement.

\subsection{Ablation Study of CAMP}

All ablations use the simulated Ball-in-Box Pregrasp task and 100 planning
trials per setting. All other planning and validation settings remain fixed.

\begingroup
\setlength{\parskip}{0pt}
\makeatletter
\renewcommand{\subsubsection}{\@startsection{subsubsection}{3}{\parindent}{0pt}{0pt}{\normalfont\normalsize\itshape}}
\makeatother
\subsubsection{Comparison of Fixed-Arm and Arm-Relaxed Initialization}

We compare CAMP with a variant that disables arm relaxation during local replanning. Table~\ref{tab:vmp_ablation} shows that CAMP achieves a higher S.R. Each trial performs eight initialization attempts. For the variant, 12 of the 100 trials produce no more than two valid initial trajectories, and only one of these 12 trials eventually succeeds. In contrast, CAMP produces at least two valid initial trajectories in every trial. A fixed arm guide determines the palm position and orientation along the path, restricting the hand motions available for collision-free connections between adjacent layers. Therefore, local arm relaxation can help generate complete arm--hand candidates for subsequent optimization.

\subsubsection{Comparison of Trajectory Representations}

Under our setup, waypoint-wise optimization uses 4,356 variables after
fixing the endpoints, whereas VMP uses only 500 weights while retaining
independent trajectory parameters for all active joints. To isolate the effect
of trajectory representation, we replace VMP with a waypoint-wise
parameterization and keep all other settings fixed. Table~\ref{tab:vmp_ablation}
shows that VMP increases S.R., reduces P.T., and yields a shorter mean N.T.L.
This result supports the idea that the compact representation improves optimization
efficiency without removing the continuous arm motion and hand-shape changes
required in constrained environments.

\subsubsection{Comparison of Coarse-to-Fine and Fine-Only Optimization}

To evaluate coarse-to-fine optimization, we compare the full method with a
fine-only variant that directly optimizes the initialized VMP candidates in
parallel.
The coarse stage penalizes deviations from the initial arm weights more
strongly, favoring hand-shape corrections while discouraging large changes to
the arm guide. The fine stage then reduces this penalty and increases the
relative weight of trajectory quality. Direct fine optimization produces
larger simultaneous changes in the arm route and hand shape from the outset,
making local constraint satisfaction more difficult.
\par\endgroup

\begin{figure*}[!t]
    \centering
    \setlength{\abovecaptionskip}{2pt}
    \includegraphics[width=\textwidth]{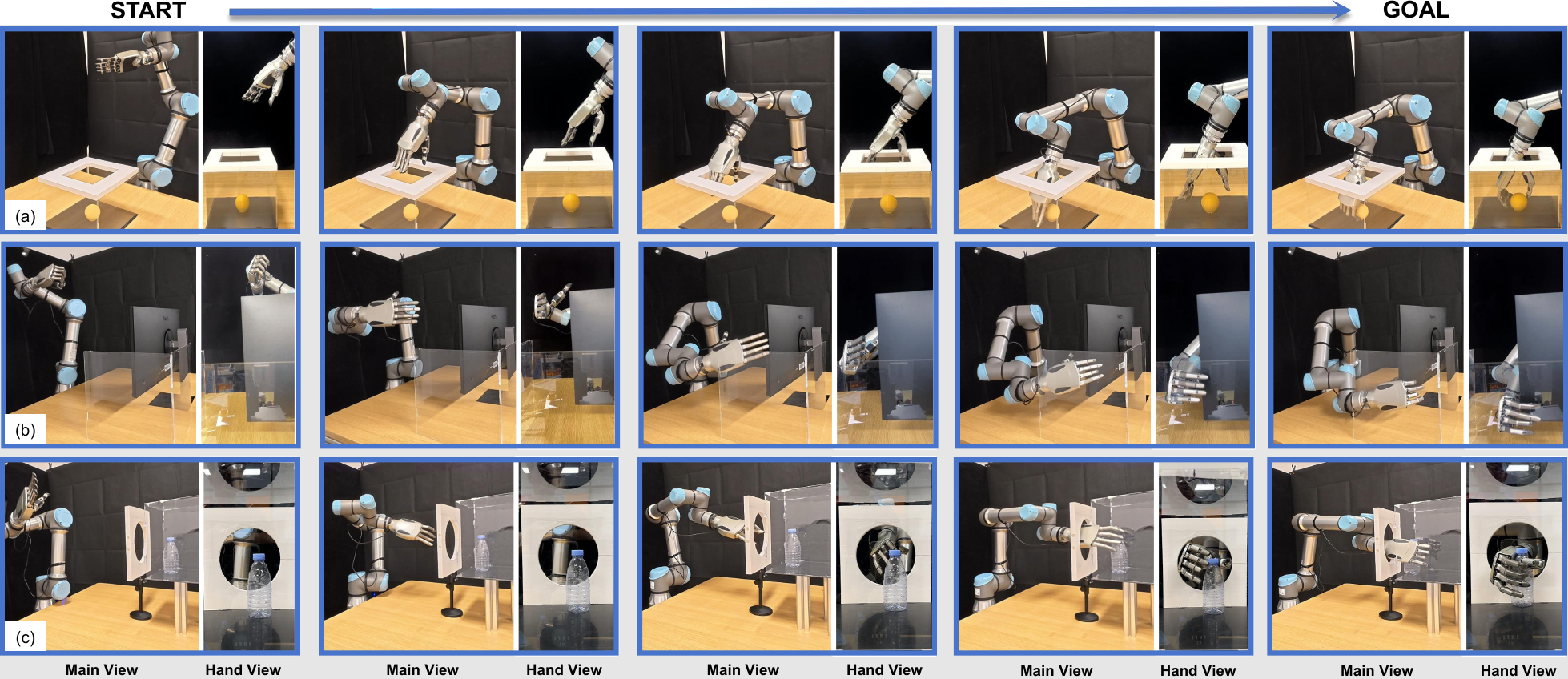}\par
    \nointerlineskip
    \caption{Real robot experiment snapshots. Time advances from left to right.
    Rows (a)--(c) show Ball-in-Box Pregrasp, Display Button Press, and Cabinet
    Cylinder Pregrasp, respectively. Secondary views show hand details from
    another viewpoint as the robot approaches the constrained regions.}
    \label{fig:real_robot_experiments}
\end{figure*}

\subsection{Real Robot Experiments}

We evaluate CAMP on the physical robot system. Each obstacle is represented by the known geometric model used in simulation, and we adjust its position for each trial. To account for obstacle-modeling errors in the physical scene, we use a 1~cm environment-collision margin.
We conduct 10 complete system trials for each of Ball-in-Box Pregrasp, Display Button Press, and Cabinet Cylinder Pregrasp, obtaining task success rates of 80\%, 90\%, and 80\%, respectively. For Ball-in-Box Pregrasp and Cabinet Cylinder Pregrasp, success requires reaching the goal pose but does not require grasping or lifting the object. The grasping and lifting trajectories are manually adjusted solely for demonstration.
For Display Button Press, success requires the fingertip to reach the prescribed target position. No physical button press is performed. Most failures involve collisions between the hand and nearby obstacles while traversing constrained regions. These failures indicate that execution in tight clearances remains sensitive to obstacle localization, camera calibration, and robot execution errors.

%% file: Sections/7_conclusion.tex
\section{Conclusions}

This paper presents CAMP for cooperative arm--hand motion planning in constrained environments. CAMP combines arm-space guidance, layered hand search with local arm relaxation, and coarse-to-fine joint optimization using an endpoint-preserving VMP representation. Across six simulated tasks, CAMP achieves mean planning success rates of 84.2\%--98.5\%. Real-robot experiments validate the feasibility of the planned motions on UR7e+LinkerHand. The current evaluation is limited to one platform and static environments with known obstacle geometry. Failures during physical execution highlight sensitivity to obstacle localization, calibration, and robot execution errors in tight clearances. Future work will incorporate modeling and execution uncertainty into collision constraints.